\documentclass[runningheads]{llncs}
\usepackage{xcolor}
\usepackage{amsmath}
\usepackage{amssymb}
\usepackage[T1]{fontenc}
\usepackage{graphicx,verbatim}
\usepackage{booktabs}

\usepackage{tikz}
\usetikzlibrary{arrows.meta}
\usepackage{xcolor}

\definecolor{mygreen}{RGB}{37,79,34}
\definecolor{myblue}{RGB}{0,11,88}

\DeclareRobustCommand{\greensample}{%
\tikz[baseline=-0.6ex]{%
\draw[mygreen, thick] (0,0.1ex) -- (0.32,0.1ex);%
}}

\DeclareRobustCommand{\bluesample}{%
\tikz[baseline=-0.6ex]{%
\draw[myblue, line width=1.2pt, dash pattern=on 2pt off 1.6pt] (0,0.1ex) -- (0.35,0.1ex);%
}}

\usepackage[table]{xcolor}
\definecolor{lightgray}{gray}{0.94}

\usepackage[colorlinks=true, 
     linkcolor=blue, 
     filecolor=blue, 
     citecolor = blue,       
     urlcolor=blue]{hyperref}
     
\begin{document}
\title{Set-Inclusive Uncertainty Modeling \\for Robust Brain Tumor Segmentation}

\author{
Seunghun Baek$^\ast$
\and
Jihwan Park$^\ast$
\and
Jaeyoon Sim
\and
Hoseok Lee,
\\
Seungjoo Lee
\and
Won Hwa Kim
} 
\authorrunning{S. Baek \& J. Park et al.}
\titlerunning{Set-Inclusive Uncertainty Modeling}

\institute{Pohang University of Science and Technology, Pohang, South Korea \\
\email{\{habaek4, pjh58110, simjy98, hslee0608, seungjoo612, wonhwa\}@postech.ac.kr}}

\def\thefootnote{*}\footnotetext{S. Baek and J. Park contributed equally to this paper.}
  
\maketitle              
\begin{abstract}
Multimodal MRI is essential for accurate brain tumor segmentation.
However, acquiring all modalities at inference is often challenging in practice, 
which causes intrinsic uncertainty due to unavoidable information loss. 
Without modeling this uncertainty, existing methods encode incomplete evidence into deterministic representations that appear plausible but lack reliability.
In this regime, we propose a probabilistic representation framework that models representations as Gaussian distributions,
where their mean captures task information and their variance measures uncertainty from missing evidence.
To make variance reflect information deficiency, 
we regularize the mean from each partial configuration toward its full-modality counterpart, 
while scaling the variance with the discrepancy between their aligned means.
We further introduce a set-inclusive strategy that exploits the hierarchical structure of modality subsets and enforces an ordering constraint to maintain their consistent uncertainty relationships.
Extensive experiments on BraTS 2018 and 2020 demonstrate that our approach offers superior performance over baselines across diverse missing-modality scenarios.
Code and model checkpoint are available at https://github.com/atlas-sky/SIUM.


\end{abstract}
\section{Introduction}
\label{sec:intro}
Brain tumor segmentation is critical for assessing disease progression and treatment planning~\cite{intro_1,intro_2,brats}.
In clinical practice,
four standard magnetic resonance imaging (MRI) modalities are utilized: T1, T1c, T2, and FLAIR~\cite{intro_6,intro_3}, 
which 
provide complementary information for delineating tumor subregions,
including whole tumor (WT), tumor core (TC), and enhancing tumor (ET)~\cite{intro_5,intro_4,intro_3,brats}.
Since no single modality fully captures every tumor subregion,
accurate delineation requires the joint integration of all modalities~\cite{intro_8,intro_7}.
However, 
acquiring all of them is often impractical due to cost and patient burden~\cite{intro_10,intro_9}.
As brain tumor segmentation relies on the complementary evidence from multiple modalities, 
when a modality goes missing,
its unique information becomes inaccessible.

To address potential incompleteness, 
previous works~\cite{rfnet,dcseg,m3ae,mmformer} train their frameworks to be robust across missing-modality scenarios.
They encode each modality configuration into a deterministic embedding to represent task-relevant information.
While the absence introduces unavoidable information loss in the resulting embedding,
deterministic approaches still update their prediction models without considering informational incompleteness (i.e., uncertainty).
Consequently, the models may rely on hallucinated features that lack evidence.

In this regime, we model each configuration as a Gaussian distribution, where the mean encodes task-relevant information and the variance captures uncertainty.
This design acknowledges that missing modalities cause intrinsic information deficiency, 
which should be reflected as uncertainty rather than concealed within a deterministic embedding. 
To make the variance encode uncertainty, we exploit the hierarchical structure of modality subsets, 
where smaller sets are nested within larger ones. 
Within this set-inclusive hierarchy, the variance is encouraged to increase when the mean of a subset deviates from its full-modality counterpart, 
since deviation indicates missing evidence. 
It is further regularized to remain higher for subsets than for their supersets, which reflects their reduced informational completeness. 
Through uncertainty guidance, the model enables reliable predictions that prioritize well-supported evidence across configurations.

\textbf{Our main contributions} are summarized as follows: 
\textbf{1)} We propose a probabilistic framework that integrates set-inclusive hierarchy with uncertainty guidance to model intrinsic uncertainty and emphasize reliable cues.
\textbf{2)} We provide a theoretical analysis of how uncertainty influences task-representation optimization and empirical validation of the learned uncertainty to support the need for probabilistic embeddings.
\textbf{3)} Extensive experiments on BraTS 2018 and 2020 demonstrate superior performance under diverse missing-modality scenarios.

\textbf{Related works on Brain Tumor Segmentation.}
Early approaches adopt imputation methods that synthesize absent modalities \cite{imputation_1,imputation_2,imputation_3}.
However, these methods require additional generative models, 
which are computationally heavy and time-consuming. 
Recent works therefore have shifted toward imputation-free strategies. 
In particular, RFNet~\cite{rfnet} introduces a region-aware fusion module that assigns adaptive modality importance across tumor regions. 
mmFormer~\cite{mmformer} improves it using inter- and intra-modal Transformer~\cite{transformer} to model local and global context.
M$^3$AE~\cite{m3ae} mitigates missing-modality effects by modality- and patch-level masking and self-distillation.
DC-Seg~\cite{dcseg} disentangles modality-invariant and -specific representations via contrastive learning for anatomical consistency. 

\section{Methods}
\label{sec:method}
\textbf{General problem setting. }
Given a sample $x_i$ (i.e., subject) with $M$ modalities, 
we denote its $m$-th modality input (e.g., T1 scan) as $x_i^{(m)}$ for $m=$$1,\dots,M$.
Each modality input $x_i^{(m)}$ is encoded by a modality-specific encoder $\mathcal{E}^{(m)}$ to produce an embedding $e_i^{(m)}=\mathcal{E}^{(m)}(x_i^{(m)})$.
To simulate missing-modality scenarios during training, a Bernoulli indicator $\delta_i^{(m)}$$\in\{0,1\}$ is adopted to mask the modality embedding, resulting in $\delta_i^{(m)}e_i^{(m)}$.
A shared decoder $\mathcal{D}$ then aggregates the set of masked embeddings $\{\delta_i^{(m)}e_i^{(m)}\}_{m=1}^{M}$ to obtain a task-specific representation
$z_i=\mathcal{D}(\{\delta_i^{(m)}e_i^{(m)}\}_{m=1}^{M})$.
While incomplete modalities inherently imply a high degree of uncertainty, existing methods~\cite{rfnet,dcseg,m3ae,mmformer} typically maintain $z_i$ as a deterministic embedding.
By directly passing $z_i$ to a task head $h$ to obtain the prediction $\hat{y}_i = h(z_i)$, such approaches do not explicitly account for the uncertainty induced by unobserved modalities.
As a consequence, the model may be trained with excessive confidence in the presence of missing modalities.

\begin{figure}[t!]
    \centering
    \includegraphics[width=0.95\linewidth]{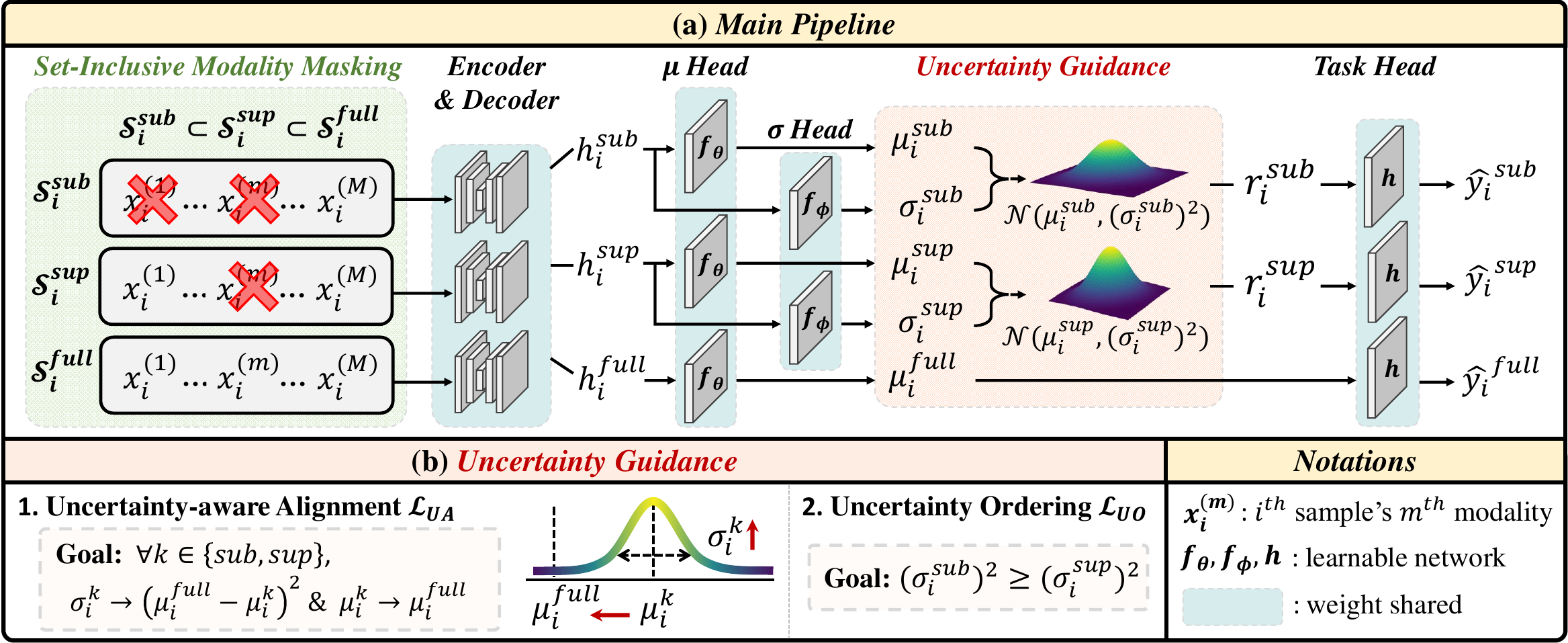}
    \caption{Overall framework.
    \textbf{(a)} For each sample $x_i$, three modality sets 
    $\mathcal{S}_i^\text{sub}$$\subset$ $\mathcal{S}_i^\text{sup}$$\subset$ $ \mathcal{S}_i^\text{full}$
    are constructed and then processed independently.
    Subset configurations are modeled probabilistically as 
    $r_i^k$$\sim$$\mathcal{N}(\mu_i^k, (\sigma_i^k)^2)$,
    whereas the full configuration serves as a deterministic anchor as $\mu_i^\text{full}$.
    \textbf{(b)}
    $\mathcal{L}_\text{UA}$ aligns each subset mean $\mu_i^k$ with the full-modality anchor while enabling $\sigma_i^k$ to reflect their discrepancy.
    $\mathcal{L}_\text{UO}$ further enforces an ordering constraint such that smaller subsets maintain higher uncertainty than their supersets.
    }

    \label{fig:framework}
\end{figure}

\subsection{Theoretical Analysis of Probabilistic Representation}
\label{unc}
To explicitly model the uncertainty,
we represent the task embedding $z_i$ in a probabilistic manner.
The task representation $z_i$ is transformed into a Gaussian embedding composed of a mean $\mu_i = f_\theta(z_i)$ and a standard deviation $\sigma_i = f_\phi(z_i)$, 
where $f_\theta$ and $f_\phi$ are learnable networks parametrized by $\theta$ and $\phi$.
The role of $\mu_i$ is to encode the task-specific embedding, 
while $\sigma_i$ models the uncertainty inherent in $\mu_i$.
We sample $r_i$ from 
$\mathcal{N}(\mu_i, \sigma_i^2)$ to obtain a probabilistic representation that captures both the embedding and its uncertainty.
To enable gradient optimization, we employ the reparameterization trick~\cite{reparameterization} on $r_i$ as
\begin{equation}\label{eq:sampling}
    r_i = \mu_i + \epsilon \odot \sigma_i, \quad \epsilon \sim \mathcal{N}(0, I).
\end{equation}
During training, task prediction is obtained as $\hat{y}_i = h(r_i)$ to incorporate uncertainty.
At inference, only $\mu_i$ is used as $\hat{y}_i = h(\mu_i)$ for the consistent prediction.

Our training introduces stochastic perturbations around $\mu_i$, 
whose magnitude is governed by $\sigma_i$, thereby influencing the loss landscape. 
To clarify this effect, we analyze how $\sigma_i$ affects the optimization of $\theta$, where $\mu_i$$=$$f_\theta(z_i)$, by examining the gradient of the loss $\mathcal{L}$ under sampling noise $\epsilon$$\sim$$\mathcal{N}(0, I)$ (i.e., $\nabla_{\theta} L(\theta;\epsilon)$).
Applying the chain rule, gradient with respect to $\theta$ decomposes into $\nabla_{r_i} L(\theta;\epsilon)$ and $\frac{\partial {r_i}}{\partial \theta}$ as
\begin{equation}\label{eq:chainrule}
\nabla_{\theta} L(\theta;\epsilon) 
= (\frac{\partial r_i}{\partial {\mu_i}}\frac{\partial {\mu_i}}{\partial \theta})^{\top} \nabla_{r_i} L(\theta;\epsilon)
= \frac{\partial {\mu_i}}{\partial \theta}^{\top} \nabla_{r_i} L(\theta;\epsilon)\;(\because \frac{\partial {r_i}}{\partial \mu_i}=\textbf{I}).
\end{equation}
Since the noise $\epsilon \odot \sigma_i$ is added to $\mu_i$ in $r_i$,
the gradient $\nabla_{r_i} L(\theta;\epsilon)$ can be linearized around the reference point $\mu_i$ (i.e., $\epsilon=0$).
A first-order Taylor expansion yields
\begin{equation}\label{eq:taylor}
\nabla_{r_i} L(\theta;\epsilon) \approx \nabla_{r_i} L(\theta;0)+\nabla_{r_i}^{2}L(\theta;0) (\epsilon \odot \sigma_i).
\end{equation}
To characterize the influence of $\epsilon$ on $\nabla_{\theta} L(\theta;\epsilon)$, 
we first analyze the covariance of $\nabla_{r_i} L(\theta;\epsilon)$ with respect to the sampling noise $\epsilon$ (i.e., $\mathrm{Cov}_\epsilon[\nabla_{r_i} L(\theta;\epsilon)]$).
Using the approximation in Eq.~\eqref{eq:taylor} and the covariance identity $\mathrm{Cov}[\mathbf{A}x]$$=$$\mathbf{A}\mathrm{Cov}[x]\mathbf{A}^{\top}$, 
the resulting gradient covariance with respect to $\epsilon$ is proportional to $\sigma_i^2$ as
\begin{equation}\label{eq:cov_s}
\begin{aligned}
\mathrm{Cov}_{{\epsilon}} [\nabla_{r_i} L(\theta;\epsilon)] 
&\approx 
\mathrm{Cov}_{{\epsilon}} [\nabla_{r_i} L(\theta;0)]+\mathrm{Cov}_{{\epsilon}} [\nabla_{r_i}^{2}L(\theta;0)(\epsilon \odot \sigma_i)]\\
&=\nabla_{r_i}^{2}L(\theta;0)\, \mathrm{diag}\!\big(\sigma_i^2\big) \nabla_{r_i}^{2}L(\theta;0)^{\top}\; (\because \mathrm{Cov}_{{\epsilon}} [{\epsilon \odot \sigma_i}]=\mathrm{diag}\!\big(\sigma_i^2\big)).
\end{aligned}
\end{equation}
The chain rule (Eq.~\eqref{eq:chainrule}) further links $\mathrm{Cov}_{\epsilon} [\nabla_{\theta} L(\theta;\epsilon)]$ to $\mathrm{Cov}_{\epsilon} [\nabla_{r_i} L(\theta;\epsilon)]$ as
\begin{equation}
\mathrm{Cov}_{\epsilon}[\nabla_{\theta} L(\theta;\epsilon)]
= \frac{\partial {\mu_i}}{\partial \theta}^{\top}\mathrm{Cov}_{\epsilon} [\nabla_{r_i} L(\theta;\epsilon)]\ \frac{\partial {\mu_i}}{\partial \theta},
\end{equation}
which also scales with $\sigma_i^2$ according to Eq.~\eqref{eq:cov_s}. 
Larger $\sigma_i$ induces higher gradient variance in the $\mu_i$-determining parameters $\theta$. 
The resulting stochastic updates fluctuate across iterations and partially cancel in expectation,
which attenuate effective parameter updates and thereby diminish their impact on the task.

\subsection{Set-Inclusive Modality Masking and Uncertainty Guidance}
\label{sec:uar}
While $\sigma_i$ modulates the updates of the model parameter $\theta$,
optimizing the task loss alone (i.e, $\mathcal{L}=\mathcal{L}_\mathrm{Task}$) collapses $\sigma_i$ toward zero.
To be explicit, analogous to the local expansion in Eq.~\eqref{eq:taylor}, 
the expected loss admits the approximation as
\begin{equation}\label{eq:sigma_collapse}
\mathbb{E}_{\epsilon}\!\left[L(\mu_i+\epsilon \odot \sigma_i)\right]
\;\approx\;
L(\mu_i)
+ \tfrac{1}{2}\,\mathrm{Tr}\!\left(\nabla^2_{r_i}L(\theta;0)\,\mathrm{diag}(\sigma_i^2)\right).
\end{equation}
Since the second term is non-negative and strictly increasing in $\sigma_i^2$,
minimizing the task loss drives $\sigma_i \to 0$,
which necessitates additional regularization to preserve stochasticity in $r_i$.
In this regime, we posit that {\it $\sigma_i$ should increase when task-relevant information is insufficient},
which naturally occurs {\it as available modalities are progressively removed.}
This motivates our subsequent design,
which incorporates set-inclusive training and uncertainty regularization.

\noindent\textbf{Set-Inclusive Modality Masking.}
Let $\mathcal{M} = \{1, \dots, M\}$ denote the index set of all modalities.
For a given sample $x_i$, each modality configuration is represented as a subset
$\mathcal{S}_i \subseteq \mathcal{M}$ indicating the observed modalities.
The masking indicator $\delta_i^{(m)}$ is set to 1 if and only if $m \in \mathcal{S}_i$,
and to 0 otherwise.
To enable set-inclusive uncertainty modeling, we consider three modality index sets
associated with the same sample $x_i$.
Let $\mathcal{S}_i^{\mathrm{full}} = \mathcal{M}$ denote the full-modality configuration.
We further define two partial modality sets $\mathcal{S}_i^{k} \in \{\mathcal{S}_i^{\mathrm{sub}}, \mathcal{S}_i^{\mathrm{sup}}\}$,
which satisfy the inclusion relation as 
$\mathcal{S}_i^{\mathrm{sub}} \subset \mathcal{S}_i^{\mathrm{sup}} \subset \mathcal{S}_i^{\mathrm{full}}$.
For each partial modality set $\mathcal{S}_i^{k}$, the task representation $z_i^{k}$
is parameterized as a Gaussian embedding with mean $\mu_i^{k}$ and variance $(\sigma_i^{k})^2$.
We 
then perform stochastic sampling (Eq.~\eqref{eq:sampling}) as $r_i^{k} = \mu_i^{k} + \epsilon \odot \sigma_i^{k}$
to obtain the task prediction $\hat{y}_i^{k}=h(r_i^{k})$.
For a full-modality set, which serves as an uncertainty-free anchor, 
the prediction $\hat{y}_i^{\mathrm{full}}$ is obtained directly from $\mu_i^{\mathrm{full}}$.
Accordingly, the task loss is defined over the modality sets as
\begin{equation}
\mathcal{L}_{\mathrm{Task}} = \ell(\hat{y}_i^{\mathrm{sub}}, y_i)+\ell(\hat{y}_i^{\mathrm{sup}}, y_i)+\ell(\hat{y}_i^{\mathrm{full}}, y_i),
\end{equation}
where $\ell$ is a task-specific loss (e.g., Dice loss)
and $y_i$ is the ground-truth label.

\sloppy
\noindent\textbf{Set-Inclusive Uncertainty Guidance.} As the full-modality configuration $S_i^\mathrm{full}$ provides the most complete evidence,
we treat its mean $\mu_i^{\mathrm{full}}$ as an uncertainty-free anchor.
For each incomplete configuration $k\in \{\mathrm{sub},\mathrm{sup}\}$, we align its mean $\mu_i^{k}$ with this anchor,
while allowing $\sigma_i^{k}$ to quantify the discrepancy between $\mu_i^{k}$ and $\mu_i^{\mathrm{full}}$.
We define the corresponding uncertainty-aware alignment loss $\mathcal{L}_{\mathrm{UA}}$
as a Gaussian negative log-likelihood with gradients detached from $\mu_i^{\mathrm{full}}$.
For $\mu_i^k,\sigma_i^k\in\mathbb{R}^{C\!\times\!H\!\times\!W\!\times\!D}$ with channel dimension $C$ and spatial shape $H$$\times$$W$$\times$$D$,
$\mathcal{L}_{\mathrm{UA}}$ is computed element-wise and averaged over all elements via $\mathbb{E}$ as
 
\begin{equation}
\mathcal{L}_{\text{UA}} = \mathbb{E}\left[\textstyle\sum_{k \in \{\mathrm{sub}, \mathrm{sup}\}} \left( \frac{1}{2} \frac{(\mu_{i}^{\text{k}} - \text{GradStop}(\mu_{i}^{\text{full}}))^2}{(\sigma_{i}^{\text{k}})^2} + \frac{1}{2} \log (\sigma_{i}^{\text{k}})^2\right)\right].
\end{equation}
The former term penalizes deviations of $\mu_i^{k}$ from $\mu_i^{\mathrm{full}}$,
while enlarging $\sigma_i^{k}$ when the discrepancy is substantial.
The latter prevents $\sigma_i^{k}$ from growing unbounded.
These terms
drive
$\mu_i^{k}$ toward $\mu_i^{\mathrm{full}}$ while guiding $\sigma_i^{k}$
to reflect their discrepancy.
We further impose an ordering constraint consistent with the set-inclusion hierarchy.
Since $\mathcal{S}_i^{\mathrm{sub}} \subset \mathcal{S}_i^{\mathrm{sup}}$, 
uncertainty from $\mathcal{S}_i^{\mathrm{sub}}$ should be greater.
To penalize violations of this hierarchy,
we introduce an uncertainty ordering loss $\mathcal{L}_{\mathrm{UO}}$ as
\begin{equation}
\mathcal{L}_{\mathrm{UO}} = \mathbb{E} \left[ \max \bigl(0, (\sigma_i^{\mathrm{sup}})^2 - (\sigma_i^{\mathrm{sub}})^2 \bigr) \right].
\end{equation}

The overall objective $\mathcal{L}$ combines the task loss with our two uncertainty
regularization terms,
weighted by hyperparameters $\alpha$ and $\beta$, 
respectively, as
\begin{equation}
\mathcal{L} = 
\mathcal{L}_{\text{Task}} + \alpha \mathcal{L}_{\text{UA}} + \beta \mathcal{L}_{\text{UO}}.
\end{equation}

\section{Experimental Settings}
\label{sec:exp}
\subsection{Datasets and Evaluation Metrics}
We evaluate our framework on two brain tumor segmentation benchmarks, BraTS 2018 and 2020~\cite{brats}, to assess performance across cohorts of different scales and compositions. 
Both datasets provide four MRI modalities (i.e., T1, T1c, T2, and FLAIR) and standardized tumor annotations, and performance is evaluated on WT, TC, and ET.
All volumes are skull-stripped, aligned to a common anatomical template, and resampled to an isotropic resolution of $1$mm$^3$, followed by zero-mean, unit-variance normalization within the brain region. 
BraTS 2018 contains 285 subjects, whereas BraTS 2020 extends the cohort to 369 subjects. 

We follow the same train, validation, and test splits as M$^3$AE~\cite{m3ae} for BraTS 2018 (199:29:57) and RFNet~\cite{rfnet} for BraTS 2020 (219:50:100).
During training, volumetric patches of 
$112$$\times$$112$$\times$$112$ are randomly extracted.
At inference, we adopt the sliding-window strategy of RFNet~\cite{rfnet}
and average overlapping predictions.
Performance is evaluated using the Dice similarity coefficient (DSC)~\cite{dice}.

\begin{table*}[t!]
    \centering
    \caption{Test performance measured by the Dice similarity coefficient (DSC, \%) on the BraTS 2020 dataset. 
    For each modality, $\bullet$ and $\circ$ denote its presence and absence, respectively. 
    The best and second-best results are highlighted in {\bf bold} and \underline{underline}.}
    \label{tab:brats2020}
    \resizebox{\textwidth}{!}{
    \begin{tabular}{cccc|
    cccc>{\columncolor{lightgray}}c|
    cccc>{\columncolor{lightgray}}c|
    cccc>{\columncolor{lightgray}}c}
    \toprule
    \multicolumn{4}{c|}{Modalities} & 
    \multicolumn{5}{c|}{Whole tumor (WT)} & 
    \multicolumn{5}{c|}{Tumor core (TC)} & 
    \multicolumn{5}{c}{Enhancing tumor (ET)} \\
    \midrule
    F & T1 & T1c & T2 &
    RFNet & mmFormer & M\textsuperscript{3}AE & DC-Seg & Ours &
    RFNet & mmFormer & M\textsuperscript{3}AE & DC-Seg & Ours &
    RFNet & mmFormer & M\textsuperscript{3}AE & DC-Seg & Ours \\
    \midrule
    $\circ$ & $\circ$ & $\circ$ & $\bullet$ & 86.05 & 85.51 & 86.10 & \underline{86.72} & \textbf{87.24} & 71.02 & 63.36 & \underline{71.80} & 70.88 & \textbf{73.23} & 46.29 & \underline{49.09} & 47.10 & 47.76 & \textbf{49.55} \\
    $\circ$ & $\circ$ & $\bullet$ & $\circ$ & 76.77 & 78.04 & 78.90 & \underline{79.54} & \textbf{80.11} & 81.51 & 81.51 & 83.60 & \underline{84.62} & \textbf{86.15} & 74.85 & 78.30 & 73.60 & \underline{78.90} & \textbf{81.93} \\
    $\circ$ & $\bullet$ & $\circ$ & $\circ$ & 77.16 & 76.24 & \underline{79.00} & 78.47 & \textbf{80.19} & 66.02 & 63.23 & \textbf{69.40} & \underline{66.63} & 64.88 & 37.30 & 37.62 & 40.40 & \textbf{42.19} & \underline{41.63} \\
    $\bullet$ & $\circ$ & $\circ$ & $\circ$ & 87.32 & 86.54 & \textbf{88.00} & 87.80 & \underline{87.85} & 69.19 & 64.60 & 68.70 & \underline{71.27} & \textbf{72.07} & 38.15 & 36.68 & 40.20 & \textbf{41.66} & \underline{40.56} \\
    $\circ$ & $\circ$ & $\bullet$ & $\bullet$ & 87.74 & 87.52 & 87.10 & \underline{88.17} & \textbf{88.43} & 83.45 & 82.69 & 85.60 & \underline{86.34} & \textbf{86.89} & 75.93 & 77.20 & 76.00 & \underline{80.43} & \textbf{83.31} \\
    $\circ$ & $\bullet$ & $\bullet$ & $\circ$ & 81.12 & 80.70 & 80.10 & \underline{82.22} & \textbf{83.41} & 83.40 & 82.81 & 83.80 & \underline{85.18} & \textbf{87.02} & 78.01 & \underline{81.71} & 75.30 & 79.25 & \textbf{83.11} \\
    $\bullet$ & $\bullet$ & $\circ$ & $\circ$ & 89.73 & 88.76 & 89.60 & \underline{90.01} & \textbf{90.20} & 73.07 & 71.76 & 72.80 & \underline{74.50} & \textbf{74.71} & 40.98 & 42.98 & 43.70 & \textbf{46.90} & \underline{46.81} \\
    $\circ$ & $\bullet$ & $\circ$ & $\bullet$ & 87.73 & 86.94 & 87.30 & \underline{88.09} & \textbf{88.67} & \underline{73.13} & 67.76 & 72.90 & 73.09 & \textbf{73.53} & 45.65 & \underline{49.12} & 48.70 & \textbf{50.19} & 48.86 \\
    $\bullet$ & $\circ$ & $\circ$ & $\bullet$ & 89.87 & 89.49 & 90.10 & \underline{90.32} & \textbf{90.42} & 74.14 & 70.34 & 74.30 & \underline{75.11} & \textbf{76.34} & 49.32 & 49.06 & 47.10 & \textbf{51.32} & \underline{50.82} \\
    $\bullet$ & $\circ$ & $\bullet$ & $\circ$ & 89.89 & 89.31 & 89.50 & \underline{89.99} & \textbf{90.44} & 84.65 & 83.79 & 85.50 & \underline{85.90} & \textbf{86.44} & 76.67 & 79.44 & 75.90 & \underline{80.28} & \textbf{82.25} \\
    $\bullet$ & $\bullet$ & $\bullet$ & $\circ$ & \underline{90.69} & 89.79 & 89.60 & 90.65 & \textbf{91.09} & 85.07 & 84.44 & 85.60 & \underline{86.29} & \textbf{86.40} & 76.81 & 80.65 & 76.30 & \underline{81.41} & \textbf{83.20} \\
    $\bullet$ & $\bullet$ & $\circ$ & $\bullet$ & 90.60 & 89.83 & 90.20 & \underline{90.77} & \textbf{91.00} & 75.19 & 72.42 & 74.40 & \underline{75.53} & \textbf{75.98} & 49.92 & 50.08 & 48.20 & \textbf{52.05} & \underline{50.80} \\
    $\bullet$ & $\circ$ & $\bullet$ & $\bullet$ & \underline{90.68} & 90.49 & 90.50 & 90.62 & \textbf{91.27} & 84.97 & 83.94 & 85.80 & \underline{86.21} & \textbf{86.39} & 77.12 & 78.73 & 77.40 & \underline{79.42} & \textbf{81.54} \\
    $\circ$ & $\bullet$ & $\bullet$ & $\bullet$ & 88.25 & 87.64 & 87.40 & \underline{88.73} & \textbf{89.14} & 83.47 & 83.66 & 85.80 & \underline{86.49} & \textbf{86.96} & 76.99 & 77.34 & 78.00 & \underline{81.66} & \textbf{83.55} \\
    $\bullet$ & $\bullet$ & $\bullet$ & $\bullet$ & \underline{91.11} & 90.54 & 90.40 & 90.95 & \textbf{91.58} & 85.21 & 84.61 & 86.20 & \textbf{86.46} & \underline{86.38} & 78.00 & 79.92 & 77.50 & \underline{81.52} & \textbf{83.35} \\
    \midrule
    \multicolumn{4}{c|}{Average} & 86.98 & 86.49 & 86.90 & \underline{87.54} & \textbf{88.07} & 78.23 & 76.06 & 79.10 & \underline{79.63} & \textbf{80.22} & 61.47 & 63.19 & 61.70 & \underline{65.00} & \textbf{66.02} \\
    \bottomrule
    \end{tabular}
    }
\end{table*}

\begin{table*}[t]
\centering

\begin{minipage}[t]{0.46\linewidth}
\centering
\caption{Average Dice similarity coefficient (DSC, \%) on BraTS 2018 dataset.}
\label{tab:BraTS18}

\setlength{\tabcolsep}{4pt}
\renewcommand{\arraystretch}{0.965}

\resizebox{0.8\linewidth}{!}{
\begin{tabular}{c | c c c }
\hline
Methods & WT & TC & ET \\
\hline
RFNet & 85.49 & 76.00 & 58.44 \\
mmFormer$^*$ & 86.20 & 73.65 & 56.20 \\
M$^3$AE & 85.82 & 77.37  & 59.85 \\
DC-Seg$^*$ & 86.59 & 77.14 & 59.55 \\
\rowcolor{lightgray} Ours & \textbf{87.10} & \textbf{78.37} & \textbf{62.52} \\
\hline
\end{tabular}
}
\end{minipage}
\hfill
\begin{minipage}[t]{0.52\linewidth}
\centering
\caption{Ablation study on BraTS 2020.
(\textit{Set$\,\&$Prob}: set-inclusive probabilistic design)}
\label{tab:ablation}

\setlength{\tabcolsep}{3pt}
\renewcommand{\arraystretch}{0.986}

\resizebox{0.815\linewidth}{!}{
\begin{tabular}{c c c | c c c}
\hline
\textit{Set$\&$Prob}  & $\mathcal{L}_\mathrm{UA}$ & $\mathcal{L}_\mathrm{UO}$ 
& WT & TC & ET \\
\hline
$\times$ & $\times$& $\times$ & 86.98 & 78.23 & 61.47  \\
$\checkmark$ & $\times$& $\times$ & 87.54 & 79.20 & 63.74 \\
$\checkmark$ & $\checkmark$ & $\times$ & 87.73 & 79.32 & 65.57 \\
$\checkmark$ & $\times$ & $\checkmark$ & 87.57 & 79.31 & 64.61 \\
\rowcolor{lightgray}$\checkmark$ & $\checkmark$ & $\checkmark$ & \textbf{88.07} & \textbf{80.22} & \textbf{66.02} \\
\hline
\end{tabular}
}
\end{minipage}
\end{table*}

\subsection{Baselines and Implementation Details}
To validate our work, we compare our method with recent state-of-the-art approaches~\cite{rfnet,dcseg,m3ae,mmformer}. 
We report the results on each dataset from the original papers when available. 
Otherwise, we reproduce them using the official code and hyperparameters. 
These reproduced results are marked with * in Tab.~\ref{tab:BraTS18}.

Following the baselines~\cite{dcseg,m3ae,mmformer}, 
we adopt the backbone architecture and training protocol of RFNet~\cite{rfnet}. 
The output of the penultimate layer (i.e., the feature representation before the segmentation head) is replaced with a probabilistic embedding.
Two separate 3D CNN layers are used for $f_\theta$ and $f_\phi$, respectively.
Our framework was trained for 800 epochs using the Adam optimizer with a learning rate of $2$$\times$$10^{-4}$ and a batch size of 2, where $\alpha$$=$$0.01$ and $\beta$$=$$5$.

\section{Results and Analysis}
\subsection{Performance on the BraTS 2018 and 2020 datasets}

\textbf{Quantitative Results.} 
We first compare our method with recent baselines on BraTS 2020 dataset in Tab.~\ref{tab:brats2020}.
Our method outperformed the baselines by a clear margin in terms of the averaged Dice similarity score across the three tumor subregions.
In particular, it consistently achieved the best or second-best performance under almost every modality configuration, 
which demonstrates its robustness against diverse missing-modality scenarios.
A similar trend was observed on BraTS 2018 dataset, as shown in Tab.~\ref{tab:BraTS18}. 
These consistent improvements across the datasets indicate our superiority across underlying data cohorts.

\begin{figure}[t!]
    \centering
    \includegraphics[width=0.97\linewidth]{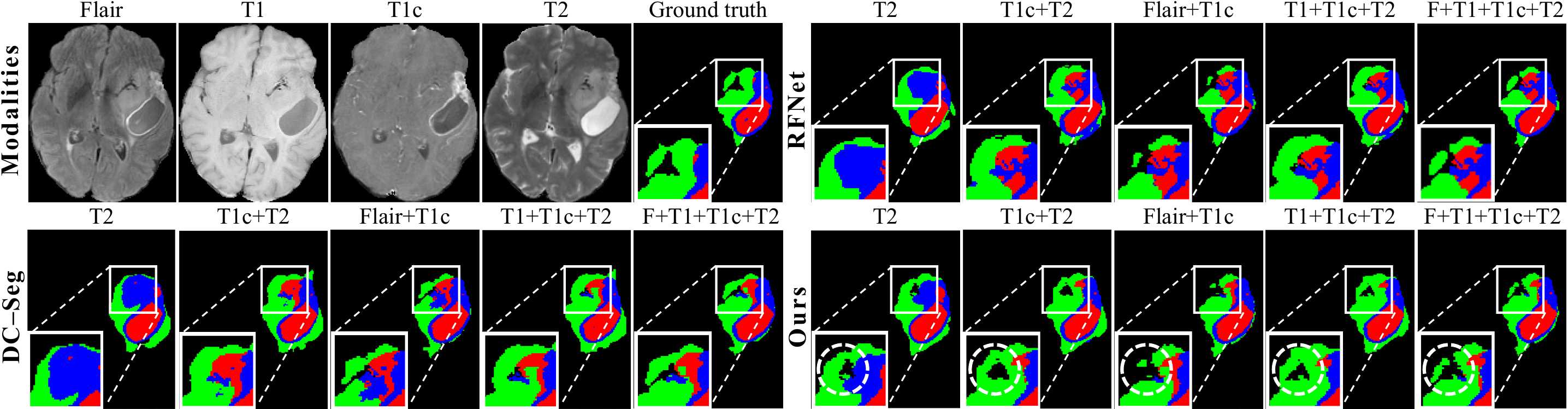}
    \caption{Qualitative results on BraTS 2020. 
    Our model produces improved segmentations over RFNet~\cite{rfnet} and DC-Seg~\cite{dcseg} across various configurations.
    (Region: 
    ET (\textcolor{blue}{blue}) $\subset$ TC (\textcolor{blue}{blue}, \textcolor{red}{red}) $\subset$ WT (\textcolor{blue}{blue}, \textcolor{red}{red}, \textcolor{green}{green}), Sample: \textit{``HG\_BraTS20\_Training\_357''})}
    \label{fig:qualitative}
\end{figure}

\begin{figure}[t!]
    \centering
    \includegraphics[width=0.91\linewidth]{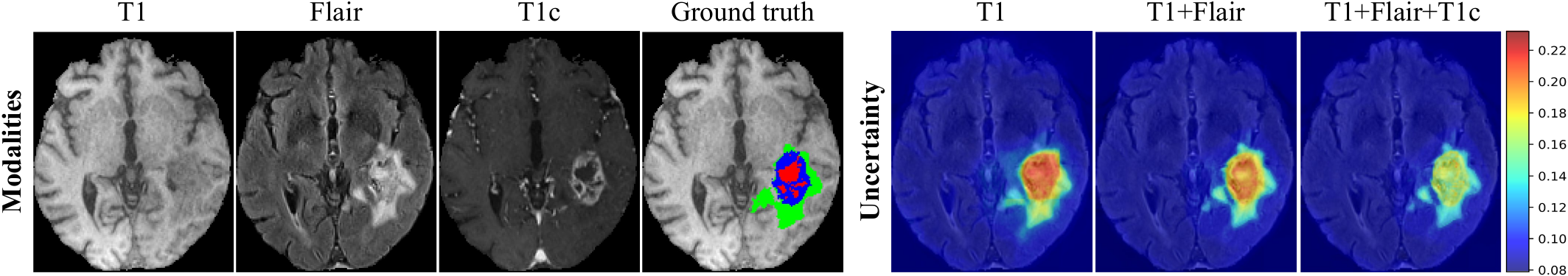}
    \caption{Visualization of input modalities and ground-truth mask, with trained uncertainty overlaid on T1. Uncertainty, initially higher within tumor regions, decreases as additional modalities are incorporated. 
    (Sample: {\it ``HG\_BraTS20\_Training\_045''})
    }
    \label{fig:sigma_map}
\end{figure}

\noindent\textbf{Qualitative Results.}
We visualize center-slice predictions on BraTS 2020 under modality configurations, 
together with RFNet~\cite{rfnet} and DC-Seg~\cite{dcseg} (Fig.~\ref{fig:qualitative}).
Our model yields more accurate and consistent segmentations across various configurations, 
notably preserving the background regions adjacent to the whole tumor.
In contrast, the baselines overfill these regions (in blue) and overestimate the ET. 
These results indicate that our model relies on reliable cues, whereas deterministic embeddings in the baselines yield spurious hallucinations.

\begin{figure}[t]
    \centering
    \includegraphics[width=0.97\linewidth]{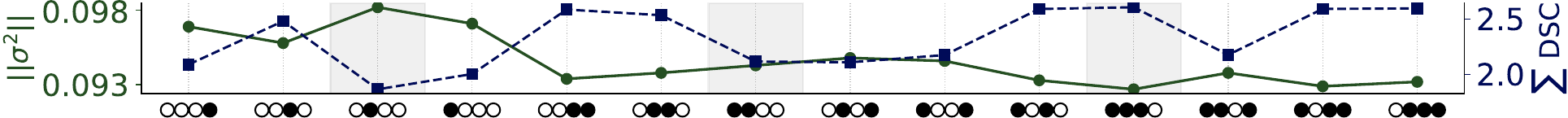}
    \caption{Variance magnitude 
    (\greensample)
    and summed test DSC across tumor subregions 
    (\bluesample)
    for each modality configuration on BraTS 2020. 
    $\bullet$ and $\circ$ indicate observed and missing modalities (Flair, T1, T1c, T2).
    Higher uncertainty correlates with lower performance.
    The shaded regions highlight that a superset exhibits lower uncertainty than its subset.
    }
    \label{fig:uncertainty}
\end{figure}

\begin{figure}[t]
    \centering
    \includegraphics[width=0.85\linewidth]{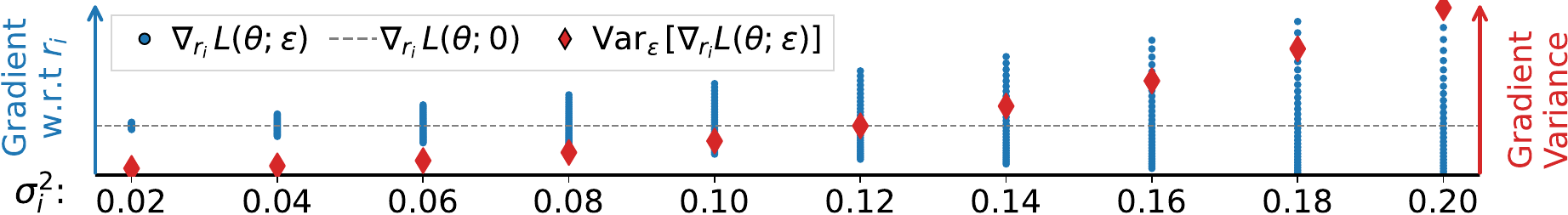}
    \caption{Effect of variance magnitude on gradient behavior under controlled scaling.
    }
    \label{fig:gradient_sigma}
\end{figure}

\subsection{Model Behavior Analyses} 

\noindent\textbf{Ablation Study.}
We analyze the contribution of each component over the deterministic baseline (i.e., RFNet~\cite{rfnet}) on BraTS 2020 (Tab.~\ref{tab:ablation}).
While adopting probabilistic embeddings with set-inclusive modality masking improves segmentation,
the variance collapse restricts its benefit (see Eq.\eqref{eq:sigma_collapse}).
$\mathcal{L}_\mathrm{UA}$ mitigates this collapse and further enhances performance, whereas $\mathcal{L}_\mathrm{UO}$ alone brings marginal gains as it constrains ordering rather than magnitude.
Combining $\mathcal{L}_\mathrm{UA}$ and $\mathcal{L}_\mathrm{UO}$ yields the best results by providing complementary uncertainty supervision.

\noindent\textbf{Uncertainty Analysis.}
We first examine the spatial distribution of learned uncertainty by averaging $\sigma$ across channels and overlaying it on the T1 image (Fig.~\ref{fig:sigma_map}).
Elevated uncertainty is primarily localized within tumor regions, particularly the enhancing tumor (blue label).
By incorporating Flair and T1c together,
uncertainty in these regions progressively decreases, indicating that complementary evidence reduces ambiguity and stabilizes model updates.

We further investigate the relationship between uncertainty and performance on BraTS 2020 across modality configurations (Fig.~\ref{fig:uncertainty}).
Uncertainty is quantified by the mean squared magnitude of $\sigma$ during training, denoted as $||\sigma^2||$.
Performance is measured by the summed Dice similarity coefficient over the three tumor subregions, denoted as $\sum \mathrm{DSC}$.
A clear inverse relationship is observed that configurations with larger variance magnitudes consistently correspond to lower test performance, and vice versa. 
This trend suggests that when missing modalities induce greater task-relevant information loss, the model responds by increasing uncertainty rather than committing to incomplete evidence.

Moreover, as highlighted by the shaded samples, superset configurations consistently exhibit lower uncertainty and higher performance than their subsets.
This ordering behavior {\em aligns with our set-inclusive design}, confirming that uncertainty systematically reflects modality-induced information incompleteness.

\noindent\textbf{Gradient Scaling Analysis.}
To empirically validate Eq.~\eqref{eq:cov_s},
which predicts that uncertainty $\sigma_i$ modulates gradient covariance,
we analyze the $\nabla_{r_i}\mathcal{L}(\theta;\epsilon)$ at an arbitrary element of $r_i$ with fixed $\mu_i$ while varying $\sigma_i$ (Fig.~\ref{fig:gradient_sigma}).
For each $\sigma_i$,
we sample $\epsilon$ 30 times to compute $\nabla_{r_i}\mathcal{L}(\theta;\epsilon)$.
Compared to deterministic gradient ($\epsilon=0$; dashed line),
the sampled ones become increasingly dispersed as $\sigma_i^2$ grows (i.e., enlarged variance).
This confirms the theoretical relationship in Eq.~\eqref{eq:cov_s}.

\section{Conclusion}
In this work, we address the intrinsic uncertainty induced by missing modalities through a probabilistic framework for incomplete multimodal brain tumor segmentation.
This design explicitly reduces reliance on spurious cues, thereby yielding more reliable predictions.
Extensive experiments on BraTS 2018 and 2020 demonstrate robustness across missing-modality scenarios.
Both theoretical and empirical analyses support the necessity of our uncertainty modeling.

\begin{credits}
\subsubsection{\ackname}
This research was supported by 
RS-2025-02216257 (60\%),
RS-2026-25494850 (30\%), 
RS-2019-II1091906 (AI Graduate Program at POSTECH, 5\%), and 
RS-2022-II220290 (5\%).

\subsubsection{\discintname}
The authors have no competing interests to declare that are relevant to the content of this article.
\end{credits}

\bibliographystyle{splncs04}
\bibliography{Paper-1877}

\end{document}